# Extracting Thyroid Nodules Characteristics from Ultrasound Reports Using Transformer-based Natural Language Processing Methods

**Aman Pathak, MS[1], Zehao Yu, MS[1], Daniel Paredes, MS[1], Elio Paul Monsour, MD[2], Andrea Ortiz Rocha, MD[2], Juan P. Brito, MD[3], Naykky Singh Ospina, MD[2*], Yonghui Wu, PhD[1*]**

**[1]Department of Health Outcomes and Biomedical Informatics, College of Medicine, University of Florida, Gainesville, FL, USA; [2]Division of Endocrinology, Department of Medicine, College of Medicine, University of Florida, Gainesville, FL, USA; [3]Division of Endocrinology, Diabetes, Metabolism and Nutrition, Mayo Clinic Rochester, USA.**

## Abstract

*The ultrasound characteristics of thyroid nodules guide the evaluation of thyroid cancer in patients with thyroid nodules. However, the characteristics of thyroid nodules are often documented in clinical narratives such as ultrasound reports. Previous studies have examined natural language processing (NLP) methods in extracting a limited number of characteristics (<9) using rule-based NLP systems. In this study, a multidisciplinary team of NLP experts and thyroid specialists, identified thyroid nodule characteristics that are important for clinical care, composed annotation guidelines, developed a corpus, and compared 5 state-of-the-art transformer-based NLP methods, including BERT, RoBERTa, LongFormer, DeBERTa, and GatorTron, for extraction of thyroid nodule characteristics from ultrasound reports. Our GatorTron model, a transformer-based large language model trained using over 90 billion words of text, achieved the best strict and lenient F1-score of 0.8851 and 0.9495 for the extraction of a total number of 16 thyroid nodule characteristics, and 0.9321 for linking characteristics to nodules, outperforming other clinical transformer models. To the best of our knowledge, this is the first study to systematically categorize and apply transformer-based NLP models to extract a large number of clinical relevant thyroid nodule characteristics from ultrasound reports. This study lays ground for assessing the documentation quality of thyroid ultrasound reports and examining outcomes of patients with thyroid nodules using electronic health records.*

## Introduction

Thyroid cancer overdiagnosis is common, harmful, and costly. More than 44,000 new cases of thyroid cancer are expected in the US in 2022.[1] This rate of diagnosis has increased from 5 to 15 persons per 100,000 between 1975 and 2013, the fastest increase among cancers.[2] This increase is explained, partially, by enhanced detection of small and low-risk papillary thyroid cancers that, if left undiagnosed and untreated, would pose no harm to patients.[3,4] The increased diagnosis and treatment of indolent thyroid cancer are not without consequences. Patients who receive a diagnosis of thyroid cancer usually undergo thyroid surgery, radioactive iodine therapy, and receive lifelong treatment with thyroid hormone replacement all associated with potential harm (e.g., vocal cord damage from surgery, dry mouth/sour test after iodine treatment).[5] The cost of caring for patients with thyroid cancer is expected to exceed $3.5 billion in 2030.[6]

Evaluation of patients with thyroid nodules according to thyroid cancer risk is necessary to improve the quality of care and mitigate overdiagnosis. Current clinical guidelines recommend stratification of patients according to their risk for thyroid cancer, which is mostly driven by the ultrasound characteristics of the nodule (e.g., size, presence of calcifications, composition).[4,7,8] Yet, despite the critical role that thyroid ultrasound reporting plays in the evaluation of patients with thyroid nodules, stakeholders interested in understanding the appropriateness of diagnostic investigations in patients with thyroid nodules and drivers to overdiagnosis need to spend significant resources, to manually review electronic medical records to obtain unstructured data from the thyroid ultrasound report. This laborious, expensive and time-consuming methodology limits thyroid cancer research in large databases, and the implementation of strategies to mitigate thyroid cancer overdiagnosis.[3,4,9] Natural language processing (NLP) is the key technology to extracting patient information from clinical narratives.[10] The development of an NLP pipeline for the evaluation of thyroid ultrasound reports is the first step in the development of validated and easily deployable algorithms for large-scale understanding of the practices that lead to downstream consequences of overdiagnosis and that can inform targeted interventions across healthcare systems based on their own clinical practice performance (i.e., excessive evaluation of low-risk nodules, lack of evaluation of high-risk nodules).

Previous studies have explored NLP to extract thyroid nodules characteristics from ultrasound reports. Chen *et al.*[11] examined an existing rule-based clinical NLP package clinical Text Analysis and Knowledge Extraction System (cTAKES)[12] to recognize thyroid nodules and 5 characteristics including Shape, Echogenic Foci, Echogenicity, Margins, and Composition. They developed a dataset of 153 US reports and reported that cTAKES identified 91.9% nodules and identified the gap that only very small proportion of ultrasound repots (0.7%) had enough detail to calculate a risk score category for thyroid cancer. Dedhia *et al.*[13] also examined cTAKES using a larger dataset of 243 ultrasound reports for extraction of 9 characteristics and reported an overall F1-score of 81.22% (precision of 74% and recall of 90%). They also identified that NLP systems were challenged by documenting multiple nodules (e.g., "several similar-appearing well-circumscribed slightly hypoechoic nodules"), context-dependent coreferences (e.g., "The largest nodule"). Santos *et al.*[14] applied various machine learning models including the transformer-based NLP models to predict the thyroid cancer risk categories using ultrasound reports and reported good performance. Short *et al.*[15] applied a convolutional neural network (CNN) to recognize incidental thyroid nodules on cross sectional imaging meeting criteria for sonographic follow-up. Previous studies demonstrated the feasibility of using NLP methods to extract a limited number of thyroid nodules and characteristics (<9). Until now, no studies have systematically and comprehensively extracted all clinically relevant thyroid nodule characteristics from ultrasound reports. Additionally, most of the previous studies applied rule-based systems such as cTAKES. Recently, deep learning-based NLP methods especially transformers, demonstrated advanced performance in clinical concept extraction.[16] There are limited studies exploring transformer-based NLP solutions for comprehensive extraction of thyroid nodules characteristics. In this study, a multidisciplinary team of NLP experts and thyroid specialists, identified thyroid nodule characteristics that are important for clinical care, systematically examined and categorized thyroid nodule information in US reports, composed annotation guidelines, created a corpus, and explored 5 state-of-the-art transformer-based NLP methods for extraction of 16 thyroid nodule-related characteristics from ultrasound reports.

*Corresponding author, Naykky.SinghOspina@medicine.ufl.edu, Yonghui.Wu@ufl.edu

## Methods

### Data Source

This study used EHR data from the University of Florida Health Integrated Data Repository. UF Health IDR is a clinical data warehouse that aggregates EHR data from various UF Health clinical and administrative information systems. We identified patients between 18-89 years old, undergoing thyroid surgery, biopsy, or ultrasound (using the Thyroid Ultrasound Procedure CPT Code: 76536) at UF Health between January 1, 2013 - December 31, 2020. Then, we identified a total of 184,560 clinical notes. This study was approved by the UF Institutional Review Board (IRB202201016).

### Annotation

We conveyed a multidisciplinary research team including expert NLP researchers (AP, DP, YW) and thyroid specialists (EPM, AOR, JPB, NSO) to develop annotation guidelines and monitor the annotation quality. The clinical team identified important thyroid nodules features according to clinical guidelines for the care of patients with thyroid nodules.[4,17,18] Our thyroid specialists reviewed a subset of randomly sampled notes and identified keywords "us thyroid" and "thyroid nodules" and note type "IMAGING", which were used to filter the 184,560 reports to select a total of 8,857 notes containing rich thyroid nodule characteristics. We randomly sampled 500 from the 8,857 ultrasound reports for annotation. The expert panel first composed an initial version of annotation guidelines based on domain experts' knowledge, which were iteratively optimized and updated in the training sessions and annotation rounds. Before annotation, we conducted training sessions to familiarize annotators with the guidelines (EPM, AOR) and facilitate implementation of the annotation guidelines. Three rounds of training were completed, until annotators achieved a good agreement. Then, two annotators annotated larger batches of reports independently (40 set of notes weekly). The expert panel identified the following important thyroid nodule characteristics (16 categories):

(1) Thyroid nodule: refers to various ways to identify thyroid nodules, for example, "thyroid nodule", "nodules"; including references such as *"Highest scoring TI-RADS nodule(s)", "largest nodule"*.
(2) Cervical lymph nodes: refers to the mention of cervical lymph node, for example, "*lymph nodes*".
(3) Numeric Size: refers to the size of the thyroid nodule, usually provided in 3 measurements and in centimeters. For example, "*2.2 x 1.2 x 3.2 cm", "> 1cm", "between 1-2 cm"*.
(4) Quantitative size: refers to the quantitative size of the thyroid nodule, usually provided as a verbal description. For example, "*small", "large"*.
(5) Laterality: refers to the main location of the thyroid nodule within the thyroid. For example, *"Right lower lobe", "Upper left lobe", "isthmus"*.
(6) Location: refers to the location of the thyroid nodule within main sections of the thyroid. Examples include *"upper", "middle", "lower"*.
(7) Composition: refers to the degree of solid or fluid component of the thyroid nodule. For example, *"solid", "Cystic", "purely cystic", "mostly cystic", "completely cystic", "spongiform", "mixed cystic and solid"*.
(8) Echogenicity: refers to the different refraction on ultrasound. For example, "*hyperechoic", "isoechoic", "hypoechoic", "very hypoechoic", "can't determine"*.
(9) Margins: refers to the margins/borders of the thyroid nodule. For example, *"smooth", "ill defined",*

*"lobulated", "irregular", "extra thyroidal extension", "well defined", "discrete", "poorly defined"*.

(10) Shape: refers to the shape and relation between the AP and transverse axis of the thyroid nodule. For example, *"wider than tall", "taller than wider", "oval", "round"*.

(11) Echogenic foci: refers to the presence of areas of high echogenicity. For example, *"comet tail artifact", "macrocalcification", "peripheral calcification", "rim calcification", "echogenic foci"*.

(12) Vascularity: Refers to the degree of blood flow in the thyroid nodule. For example, *"high", "normal", "low vascularity", "increase/decreased vascularity"*.

(13) TIRADS Score: summative description of thyroid cancer risk according to ACR-TIRADS. For example: "*TIR (1, 2, 3, 4, 5), TIRADS (1, 2, 4, 5), ACR-TIRADS (1,2,3,4,5)*".

(14) TIRADS risk category: refers to the categorical TIRADs categories.

(15) Total Number of nodules: refers to the number of thyroid nodules. For example, "*multiple thyroid nodules*".

(16) Risk description: Refers to the risk for malignancy assessed by the radiologist. For example, "*normal, benign appearing, suspicious, concerning, metastatic, malignant*".

We used an 'Other' category to capture "*difficult*" cases not covered by the annotation guidelines. Two annotators manually reviewed the 500 ultrasound reports to manually identify 16 categories of thyroid nodules and characteristics. We monitored the annotation agreement and periodically collected the discrepancies from annotators and instances annotated as 'Other' and discussed them in expert panel meetings. We also improved the annotation guidelines as necessary to cover new cases and solve discrepancies as needed.

### Thyroid nodules characteristics extraction

We approach the thyroid nodule characteristics extraction as a clinical concept extraction (or named-entity recognition [NER]) task and adopted the standard beginning-inside–outside (BIO) annotation format. We applied tokenization, sentence boundary detection, and BIO format transformation using a preprocessing pipeline from a previous study.[19] Specifically, we first derived distributed representations of text using various transformer models and used a soft-max layer to calculate probability scores for each BIO category. After the nodules and nodule characteristics were identified, we linked the characteristics to the nodules to form a nodule profile using relation extraction. Specifically, we generated candidate pairs of concepts where there are relations defined in the annotation guidelines and developed transformer-based classifiers to determine the relation type. The cross-entropy loss was used for fine-tuning.

### Transformer-based deep learning models

Previous studies from our group and others showed that transformer models pretrained using clinical text achieved better performance in extracting clinical concepts. We explored 5 pretrained transformers in the clinical domain, including BERT_MIMIC, RoBERTa_ MIMIC, Longformer_ MIMIC, DeBERTa_MIMIC, and GatorTron.

- **BERT_MIMIC**: BERT is the first generation of transformer model used bidirectional representations and an encoder structure. Vaswani *et al.* [20] introduced the transformer model, the Devlin *et al.* improved it with Bidirectional Encoder Representations from Transformers (BERT)[21]. We developed BERT_MIMIC model by following the same structure of BERT but pre-trained with clinical notes from the Medical Information Mart for Intensive Care (MIMIC) dataset. We adopted the BERT model implemented in Huggingface.

- **RoBERTa_MIMIC**: RoBERTa[22] was developed based on BERT by optimizing the training strategies. RoBERTa introduced new strategies including dynamic masking, full sentence sampling, large mini-batches, large byte level encoding, and removed next sentence prediction loss. In our previous study [19], we fine-tuned the general RoBERTa model using the clinical text from the MIMIC database and developed RoBERTa_MIMIC. Similarly, the implementation in Huggingface was used.

- **LongFormer_MIMIC**: LongFormer was proposed to better handle long sequence of text using sliding window and global attention mechanism, which is a better solution when the input text exceeds the limits of traditional transformer models such as BERT.

- **DeBERTa_MIMIC**: DeBERTa[23] (Decoding-enhanced BERT with disentangled attention) further improved BERT and RoBERTa using the disentangled attention mechanism and an enhanced mask decoder, which demonstrated good performance in recent studies.

- **GatorTron:** GatorTron[24] is a BERT-style large clinical language model. We pretrained GatorTron with >90 billion words of text, including >80 billion words from >290 million notes identified at the UF Health system covering patient records from 2011–2021 from over 126 clinical departments and ~50 million encounters. These clinical narratives covered healthcare settings including but not limited to inpatient, outpatient, and emergency department visits. We used the GatorTron model with 345 million parameters for this study.[24]

## Training strategies

We approached the thyroid nodule characteristics extraction as a standard NER task and developed transformer-based NLP solutions using a training, a validation, and a test set. We trained the models using the training set and monitored the performance using the validation set. The best of each transformer model was selected based on the validation performance on the validation set. To link the characteristics of interest to the thyroid nodules and cervical lymph nodes, we applied relation extraction methods. Specifically, according to our previous solutions in applying transformer models for relation extraction, we first generated candidate pairs of concepts using heuristic rules and then developed transformer-based classifiers to determine if there is a relation in between, and what's the relation type. According to our previous study, we first applied a binary classifier to determine if there are relations and then determined the relation type using heuristic rules derived from the two concepts. Our previous research showed that the binary classification strategy works well when there is only one type of relation defined between two concepts. We used the two concepts and the sentence as input to generate representations using transformer, and then used a classification layer to perform a binary classification. Similar to the thyroid nodules and nodule characteristics extraction, we used training, validation, and test sets to train the relation classifier.

## Experiment and evaluation

We reused the pretrained clinical transformer models, including RoBERTa_MIMIC and BERT_MIMIC, developed in our previous study.[19] We also explored three new clinical transformer models including LongFormer_MIMIC, DeBERTa_MIMIC, and GatorTron. Similar to BERT_MIMIC, we fine-tuned the general LongFormer model and DeBERTa model using clinical text from the MIMIC III database to derive the clinical versions (i.e., LongFormer_MIMIC and DeBERTa_MIMIC). The GatorTron model was developed by training from scratch using >90 billion words of text (including >82 billion words of de-identified clinical text from UF Health) in our previous studies.[24] We evaluated our NLP methods using both strict (i.e., exact boundary surface string match and entity type) and lenient (i.e., partial boundary match over the surface string) precision, recall, and F1-score. The evaluation scores were calculated with the evaluation script from a pipeline implemented in a previous study.[19]

## Results

The final dataset contains a total of 490 reports, where 10 reports were removed due to duplication or contained very few thyroid information. The final annotation guide defined 16 categories of thyroid nodules and cervical lymph node characteristics and 27 types of relations between the concepts. Two annotators (EPM, AOR) manually annotated a total of 7,958 concepts and 4,762 relations from the 490 ultrasound reports. **Figure 1** shows an example for the thyroid nodule US annotation. The annotation agreement measured by Kappa scores increased from 0.45 to 0.89 after three rounds of annotation. We randomly divided the 490 notes in a ratio of 7:1:2 into the training set of 343 notes, a validation set of 49 notes to develop various transformer-based models, and a test set of 98 notes for evaluating model performance.

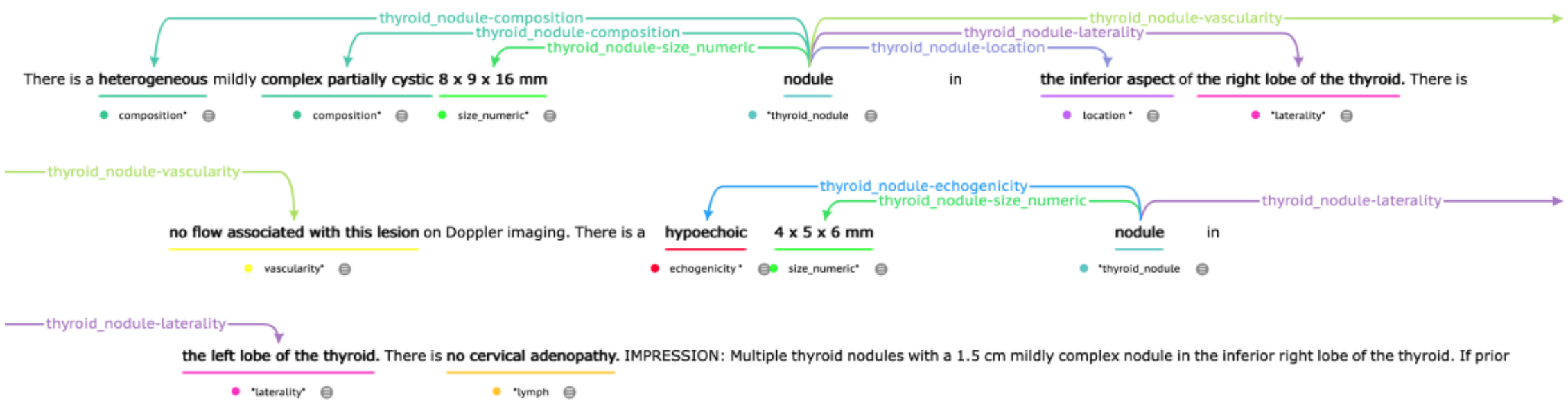

**Figure 1.** An example of thyroid nodule characteristics annotation.

**Table 1**. Comparison of transformer models for extraction of thyroid nodules and nodule characteristics.

| | Strict | | | Lenient | | |
|---|---|---|---|---|---|---|
| **Model** | **Precision** | **Recall** | **F1** | **Precision** | **Recall** | **F1** |
| BERT_MIMIC | 0.8447 | 0.8980 | 0.8705 | 0.9201 | 0.9675 | 0.9432 |
| RoBERTa _MIMIC | 0.8452 | 0.9014 | 0.8724 | 0.9185 | 0.9694 | 0.9433 |
| LongFormer_MIMIC | 0.8508 | 0.9053 | 0.8772 | 0.9207 | 0.9704 | 0.9449 |
| DeBERTa_MIMIC | 0.8564 | 0.9009 | 0.8781 | 0.9275 | 0.9699 | 0.9482 |
| GatorTron | **0.8609** | **0.9106** | **0.8851** | **0.9277** | **0.9723** | **0.9495** |

**Table 1** compares the 5 clinical transformer models using strict and lenient F1-score in extracting thyroid nodules and nodule characteristics from ultrasound reports. All clinical transformer models achieved strict F1-score over 87%, where GatorTron achieved the best F1-score of 0.8851, outperforming other clinical transformer models. For lenient F1-scores, all clinical transformers achieved comparable performance. Generally, transformer models achieved a higher recall than precision. **Table 2** shows the detailed performance for the 16 categories of thyroid nodule concepts for the GatorTron model. Among the 16 categories of concepts, GatorTron achieved strict F1-score over 90% on 6 categories, including Shape, Laterality, TIRADS_score, Size_numeric, Risk_discription, and TIRADS_risk_category. For the lenient F1-score, GatorTron achieved over 90% F1-score for 14 of the total 16 categories of concepts.

**Table 2**. Detailed performance of GatorTron model for 16 categories of thyroid nodule and characteristic concepts.

| **GatorTron** | **Strict** | | | **Lenient** | | |
|---|---|---|---|---|---|---|
| **Type** | **Precision** | **Recall** | **F1-score** | **Precision** | **Recall** | **F1-score** |
| Cervical Lymph nodes | 0.8421 | 0.8649 | 0.8533 | 0.9737 | 1.0000 | 0.9867 |
| Shape | 0.9615 | 0.8621 | 0.9091 | 0.9615 | 0.8621 | 0.9091 |
| Margins | 0.8475 | 0.8772 | 0.8621 | 0.9310 | 0.9474 | 0.9391 |
| Location | 0.8511 | 0.9302 | 0.8889 | 0.8723 | 0.9535 | 0.9111 |
| Laterality | 0.8821 | 0.9249 | 0.9030 | 0.9490 | 0.9868 | 0.9675 |
| Vascularity | 0.6977 | 0.8571 | 0.7692 | 0.8140 | 1.0000 | 0.8974 |
| Composition | 0.7337 | 0.8052 | 0.7678 | 0.9202 | 0.9740 | 0.9464 |
| TIRADS score | 0.9333 | 0.9333 | 0.9333 | 1.0000 | 1.0000 | 1.0000 |
| Size numeric | 0.9466 | 0.9696 | 0.9580 | 0.9723 | 0.9939 | 0.9830 |
| Echogenicity | 0.8932 | 0.9020 | 0.8976 | 0.9320 | 0.9412 | 0.9366 |
| Echogenic foci | 0.8421 | 0.9014 | 0.8707 | 0.9589 | 0.9859 | 0.9722 |
| Thyroid nodule | 0.7993 | 0.8918 | 0.8430 | 0.8707 | 0.9552 | 0.9110 |
| Size qualitative | 0.8393 | 0.8704 | 0.8545 | 0.9286 | 0.9630 | 0.9455 |
| Risk description | 0.9474 | 0.9474 | 0.9474 | 0.9474 | 0.9474 | 0.9474 |
| TIRADS risk category | 0.9286 | 0.9286 | 0.9286 | 1.0000 | 1.0000 | 1.0000 |
| Total number of nodules | 0.8313 | 0.8846 | 0.8571 | 0.8434 | 0.8974 | 0.8696 |
| Overall | 0.8609 | 0.9106 | 0.8851 | 0.9277 | 0.9723 | 0.9495 |

**Table 3.** Comparison of transformer models to link nodule characteristics to thyroid nodules.

| **Model** | **Precision** | **Recall** | **F1** |
|---|---|---|---|
| BERT_MIMIC | 0.8298 | 0.8922 | 0.8599 |
| RoBERTa _MIMIC | 0.8471 | 0.878 | 0.8623 |
| LongFormer_MIMIC | 0.9577 | 0.897 | 0.9264 |
| Deberta_MIMIC | 0.9348 | 0.9089 | 0.9217 |
| GatorTron_MIMIC | **0.9766** | **0.8914** | **0.9321** |

**Table 3** compares the 5 transformer models to link nodule characteristics to thyroid nodules and lymph nodes. All clinical transformer models achieved F1-score over 85%, where GatorTron achieved the best F1-score of 0.9321, outperforming other models. LongFormer_MIMIC and DeBERTa_MIMIC follows the GatorTron model, outperforming BERT_MIMIC and RoBERTa_MIMIC.

**Discussion and Conclusions**

Developing NLP systems for the evaluation of thyroid ultrasound reports is the first step to assessing the documentation quality of clinically relevant thyroid nodule characteristics and examining the outcomes of patients with thyroid nodules in large electronic health records databases. This study was led by a multidisciplinary team of NLP and thyroid specialists, that systematically and comprehensively examined thyroid nodule characteristics, composed annotation guidelines, and developed a corpus and state-of-the-art NLP models to extract thyroid nodules from ultrasound reports. We compared 5 state-of-the-art transformer models and our GatorTron model achieved the best strict and lenient F1-score of 0.8851 and 0.9495 for concept extraction, and 0.9321 for linking characteristics to nodules. Among the 16 categories of thyroid concepts, GatorTron achieved a lenient F1-score over 90% for 14 out of the 16 categories, demonstrating the feasibility of using transformer-based NLP models to extract thyroid nodules characteristics from ultrasound reports.

NLP is a branch of artificial intelligence (AI) with the potential to automate the review of large-scale patient reports, help improve accuracy and consistency of diagnosis of thyroid diseases. Researchers have developed NLP tool such as the CLAMP cancer pipeline[25] to extract tumor information from pathology reports for different types of cancers. Previous studies have explored NLP to extract a limited number of thyroid nodule characteristics (<9) from ultrasound reports. Most of the previous studies applied general purpose NLP systems that adopted rule-based solutions, such as cTAKES, it's not clear how the state-of-the-art transformer-based NLP systems could help identify the clinical important characteristics of thyroid nodules from a thyroid ultrasound report. This study systematically examined a total of 16 thyroid nodules characteristics and compared 5 state-of-the-art clinical transformer models to extend previous studies. The results of this study lay the foundation for the development of interventions to improve the care of patients with thyroid nodules. For example, clinical guidelines[26] recommend systematic description of at least 10 thyroid nodule characteristics with variable quality of reporting in routine practice that hinders the care patients with thyroid nodules receive. The NLP pipeline developed could be deployed in healthcare systems databases to better understand the frequency by which these important features are described in ultrasound reports. Similarly, the information on the ultrasound report, allows understanding of the risk for thyroid cancer in a particular thyroid nodule and subsequent clinical interventions. The NLP pipeline developed in this study, sets the foundation for evaluation of clinical outcomes for patients with thyroid nodules according to the clinical significance of their thyroid nodule. At this time, stakeholders (e.g., policy makers, researchers, clinician) interested in understanding the quality of ultrasound reporting or outcomes of patients with thyroid nodules according to ultrasound characteristics, need to manually perform extraction of the ultrasound report. Our NLP model provides an efficient alternative to facilitate this process.

Strengths of this study include the multidisciplinary collaboration between NLP and thyroid experts and identification of thyroid nodules characteristics of interest according to clinical guidelines. Similarly, UF adopted a reporting system based on the American College of Radiology guidelines for thyroid ultrasounds with a transition period ~2016, as such as cohort includes reports with different reporting strategies, likely reflective of routine clinical practice (e.g., some more systematic and adherent to guidelines than others). In addition, in this study we compared 5 state-of-the-art transformers models to extract thyroid nodule characteristics form ultrasound reports. A limitation of the study, includes training and validation of the NLP system in a single institution and requiring further studies to explore and validate our findings in different settings.

**Acknowledgement**

This study was partially supported by grants from the National Institutes of Health (NIH), National Institute on Aging, R56AG069880, a Patient-Centered Outcomes Research Institute® (PCORI®) Award (ME-2018C3-14754), and the UF Clinical and Translational Science Institute (UL1TR001427). This project was supported by the UF Informatics Institute (UFII) SEED Funds. NSO was supported by the National Cancer Institute of the National Institutes of Health under Award Number K08CA248972. The content is solely the responsibility of the authors and does not necessarily represent the official views of the funding institutions. We gratefully acknowledge the support of NVIDIA Corporation and NVAITC with the donation of the GPUs used for this research.